# Connection-Coordination Rapport (CCR) Scale: A Dual-Factor Scale to Measure Human-Robot Rapport

Ting-Han Lin
*University of Chicago*
Chicago, USA
tinghan@uchicago.edu

Hannah Dinner
*University of Illinois Urbana-Champaign*
Champaign, USA
hdinner2@illinois.edu

Tsz Long Leung
*University of Chicago*
Chicago, USA
quincyleung@uchicago.edu

Bilge Mutlu
*University of Wisconsin-Madison*
Madison, USA
bilge@cs.wisc.edu

J. Gregory Trafton
*Naval Research Laboratory*
Washington, USA
greg.trafton@nrl.navy.mil

Sarah Sebo
*University of Chicago*
Chicago, USA
sarahsebo@uchicago.edu

***Abstract*—Robots, particularly in service and companionship roles, must develop positive relationships with people they interact with regularly to be successful. These positive human-robot relationships can be characterized as establishing "rapport," which indicates mutual understanding and interpersonal connection that form the groundwork for successful long-term human-robot interaction. However, the human-robot interaction research literature lacks scale instruments to assess human-robot rapport in a variety of situations. In this work, we developed the 18-item Connection-Coordination Rapport (CCR) Scale to measure human-robot rapport. We first ran Study 1 (*N* = 288) where online participants rated videos of human-robot interactions using a set of candidate items. Our Study 1 results showed the discovery of two factors in our scale, which we named "Connection" and "Coordination." We then evaluated this scale by running Study 2 (*N* = 201) where online participants rated a new set of human-robot interaction videos with our scale and an existing rapport scale from virtual agents research for comparison. We also validated our scale by replicating a prior in-person human-robot interaction study, Study 3 (*N* = 44), and found that rapport is rated significantly greater when participants interacted with a responsive robot (responsive condition) as opposed to an unresponsive robot (unresponsive condition). Results from these studies demonstrate high reliability and validity for the CCR scale, which can be used to measure rapport in both first-person and third-person perspectives. We encourage the adoption of this scale in future studies to measure rapport in a variety of human-robot interactions.**

## I. Introduction

Robots are progressively integrated into our daily lives in a wide range of roles, from providing customer services [1]–[5] to supporting children's learning [6]–[9] and to assisting older adults with daily living tasks [10]–[14]. In these roles, robots are expected to establish harmonious long-term relationships with humans. But how can we characterize such relationships and measure them? A relevant construct discussed in human-robot interaction (HRI) literature is *rapport*. Establishing rapport between humans and robots builds trust and confidence [15], facilitates effective communication [16], [17], and aids in conflict resolution [18] in human-robot teams. The emotional bonds and shared experiences formed in the process of building human-robot rapport contribute to increased user engagement with the robot [7], [8], [19], [20] and encourage further engagement with the robot even after its novelty effect wears off [21]. Thus, establishing positive human-robot rapport can facilitate harmonious long-term interaction [22], [23]. Despite rapport's significance, the HRI community is currently lacking a validated scale to measure rapport across different contexts and interactions. In this work, we seek to develop a scale that measures rapport.

TABLE I
Connection-Coordination Rapport (CCR) Scale

| Factor One: Connection | Factor Two: Coordination |
|---|---|
| Warmth | Coordination |
| Empathy | Focus |
| Friendliness | Attentiveness |
| Sympathy | Smooth flow |
| Closeness | Equal participation |
| Positivity | Engagement |
| Liking each other | |
| Enthusiasm | |
| Respect | |
| Getting along | |
| Excitement | |
| Connection | |

The finalized Connection-Coordination Rapport (CCR) Scale developed through Studies 1–3. **Question wording:** "Rate how much you think the following was present in the interaction." **A five-point scale was used:** Strongly Disagree, Disagree, Neither Agree or Disagree, Agree, Strongly Agree. **Rapport calculation:** (1) Average item ratings from the Connection factor, (2) average item ratings from the Coordination factor, and (3) average values from (1) and (2) to get one score for rapport.

An ideal rapport scale would directly measure the construct of rapport (and not an associated construct) and its development would follow traditional scale construct guidelines [24]–[27]. Previous scales that have been used in HRI to measure rapport all contain several weaknesses. For example, some HRI researchers have sought to measure rapport through related constructs, including "attention" [7], [8], "positivity" [7], [8], "coordination" [7], [8], "closeness" [2], "satisfaction" [28], "self-connection" [2], "likeability" [15], and "desire to be co-worker" [15]. Even though these constructs are likely to be positively correlated with rapport, they are different enough from rapport and thus are not direct measures of rapport.

Other HRI researchers have used scales of rapport, but these scales mostly violate scale creation guidelines in two primary ways. First, several existing rapport scales use reverse-coded items, which are best to avoid because they can increase participants' cognitive load and lead to lower response accuracy [29]–[31]. Second, many current rapport scales include narrowly scoped items that are context-specific, which reduce the usability of the scale [26]. It is best to construct scale items that are as general as possible so they can be applied in many different situations. Additionally, and specific to rapport, all current rapport scales measure rapport *only from a first-person perspective*. However, rapport is a construct that can also be perceived by an observer who does not necessarily need to be part of the interaction (from a third-person perspective). None of the current rapport scales considered this aspect of rapport in their scale design.

The most cited rapport scale in the HRI literature is Rapport Scale 1 from Gratch et al. [32], which was developed to measure rapport between human and virtual agents. Rapport Scale 1 and its latest variant Rapport Scale 4 from Gratch et al. [33] have been adopted by multiple HRI studies [15], [18], [34]–[36], but they all contain reverse-coded items that reduce the quality of responses. For example, Rapport Scale 4 from Gratch et al. [33] contains reverse-coded items such as "I tried to communicate coldness rather than warmth." Additionally, Rapport Scale 4 is only measured from a first-person perspective, which limits its usefulness. Finally, no clear psychometric validation was performed on this scale in Gratch et al. [33].

Many other rapport scales developed specifically for HRI are quite context-specific (e.g., [1], [19], [37], [38]), restricting their applicability. For instance, the 18-item Rapport–Expectation with a Robot Scale (RERS) from Nomura and Kanda [19] was adopted in a handful of HRI works (e.g., [39]–[41]), but it contains scale items related to hypothetical scenarios. To illustrate, RERS contains an item "If the robot has been staying with me since my birth, I will want to be together with it until my death," which only applies to robots that could offer long-term companionship. RERS and the other human-robot rapport scales also measure rapport purely from a first-person perspective. In fact, all current rapport scales in HRI and psychology literature focus on a first-person perspective [1], [19], [32], [33], [37], [42]–[49]. Given these weaknesses with previous rapport scales used in HRI, we believe that there is a need to design a psychometrically validated rapport scale with short forward-coded scale items that can be used for a variety of robots and contexts.

In this work, we present a validated scale, the Connection-Coordination Rapport (CCR) Scale, to enable the measurement of rapport in human-robot interactions. This scale contains only short phrases in its scale items and can be used to measure rapport from both the first-person and third-person perspectives in various contexts. In this paper, we first introduce how we constructed the scale items based on our definition of rapport. In Study 1, we discovered two factors in our scale via exploratory factor analysis (EFA) and named them "Connection" and "Coordination." In Study 2, we then evaluated our scale with the latest version of the most cited rapport scale in HRI literature. In Study 3, we validated our scale in an in-person HRI study. The protocols for all studies in this paper were approved by the University of Chicago's Institutional Review Board (IRB24-0884).

## II. Definition of Rapport

A fundamental step in scale development research is to offer a concrete definition of the construct that the researchers intend to measure, which subsequently guides the creation of relevant scale items [26], [27], [50]. Inspired by Tickle-Degnen and Rosenthal [51], we propose the following definition of rapport:

**A feeling of mutual understanding and interpersonal connection among individuals developed through interactions.**

This definition aims to capture the essential components that characterize rapport by focusing on the importance of mutual understanding and interpersonal connection between two or more entities and by emphasizing that rapport is developed through the interactions between individuals.

## III. Initial Scale Item Generation

To generate a pool of scale items to measure rapport, we had three coders extract terms from ten dictionary definitions of rapport [1]. We also extracted phrases from rapport definitions and rapport measures from the top 20 papers retrieved from Google Scholar with the search term "rapport" [32], [51]–[69] and the top 20 papers retrieved from Google Scholar using the search term "rapport in human robot interaction" [1], [3], [4], [7], [15]–[19], [28], [34], [37], [70]–[77].

Additionally, we conducted a pre-study (Study 0) where we collected and reviewed the general public's perceptions of rapport to both verify our definition of rapport and extract rapport-related terms. In Study 0, we asked online participants recruited from Prolific platform to provide their own definition of rapport and describe features of interactions characterized by low or high rapport. 51 participants (27 women, 23 men, 1 non-binary; ages 19 to 77 years, $M$ = 36.43 and $SD$ = 11.96) took an average of 3.52 minutes ($SD$ = 3.24 mins) to complete our study and were compensated with $1.25 USD.

From the extracted rapport-related terms from dictionary definitions, literature identified through Google Scholar searches, and general public understanding of rapport, we used the Delphi method [78] to group similar themes and generated a total of 67 candidate scale items. After multiple rounds of internal review (e.g., comparing each item to our proposed definition of rapport in Section II), we refined our list to 27 items that are positive characteristics of interaction (see Table II). The sources and reasons for keeping and rejecting each of the 67 items are detailed in Supplemental Materials Table SI.

[1]American Heritage Dictionary, The Chambers Dictionary, Collins English Dictionary, The Merriam-Webster Dictionary, Oxford English Dictionary, Cambridge Dictionary, Dictionary.com, Vocabulary.com, Google's English dictionary, and Oxford Learner's Dictionaries

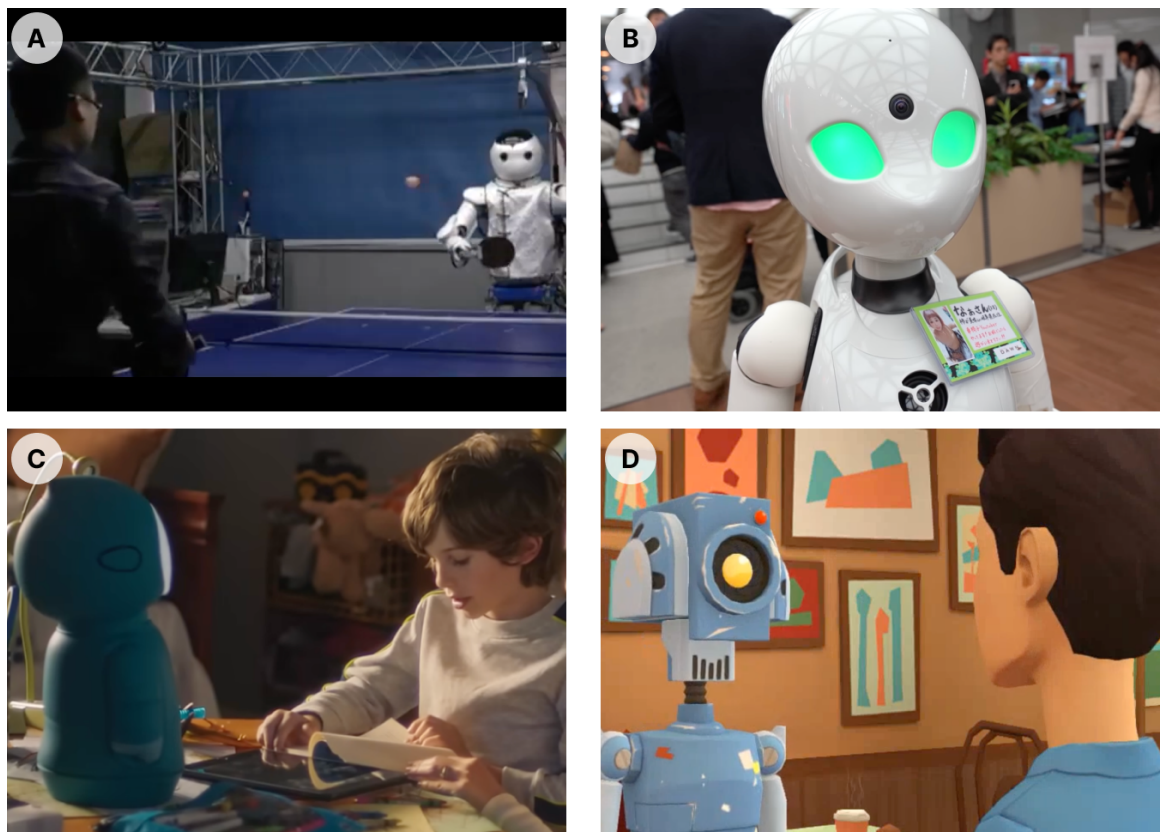

Fig. 1. Snapshots of the four videos (A) Exercise, (B) Service, (C) Companion, and (D) Argument used in Study 1.

## IV. Study 1 (Scale Construction)

Our goal in Study 1 was to refine and solidify a set of scale items generated in Section III to measure human-robot rapport.

### A. Method

We conducted an online between-subjects study asking participants to watch one video of a human-robot interaction and rate the 27 items in Table II based on the video they watched.

*1) Participants:* We recruited 300 participants who were native English speakers through the Prolific platform. The sample size of the participants was determined using the 10:1 criteria, which recommends at least 10 participants per scale item to ensure reliable validation [79]. Data from 12 participants were removed because they failed one or more attention checks in our study. From the remaining 288 participants, 204 were White, 34 were Black or African American, 18 were Asian, 16 were other ethnicity, 11 identified as two or more ethnicities, and 5 preferred not to disclose. Participants' ages ranged from 18 to 75 ($M$ = 35.92, $SD$ = 10.85). 148 participants identified as women, 133 as men, 6 as non-binary, and 1 preferred not to disclose.

*2) Materials (Videos):* We had three independent coders evaluate 30 YouTube videos featuring human-robot interactions between a single robot and a single person and place them on a linear scale from low to high rapport. From this set, we then selected four videos [80]–[83] that varied in level of rapport, context, and robot characteristics. We labeled these videos as (a) *Exercise* – a humanoid robot plays ping pong with a human, (b) *Service* – a humanoid robot waiter converses with a human customer, (c) *Companion* – a humanoid robot encourages social-emotional learning with a child, and (d) *Argument* – a humanoid robot insults and fights with a human. Snapshots of these videos are shown in Figure 1. We trimmed these videos to be under one minute to minimize participant

TABLE II
Study 1 Exploratory Factor Analysis (EFA) Results

| Item | Connection | Coordination |
|---|---|---|
| Warmth | **1.03** | -0.19 |
| Empathy | **1.00** | -0.17 |
| Friendliness | **0.97** | -0.05 |
| Sympathy | **0.92** | -0.12 |
| Closeness | **0.85** | -0.02 |
| Positivity | **0.84** | 0.10 |
| Liking each other | **0.82** | 0.13 |
| Enthusiasm | **0.76** | 0.02 |
| Deep conversation | **0.74** | -0.22 |
| Respect | **0.74** | 0.21 |
| Getting along | **0.72** | 0.25 |
| Excitement | **0.71** | 0.06 |
| Connection | **0.63** | 0.23 |
| Coordination | -0.17 | **0.97** |
| Focus | -0.19 | **0.95** |
| Attentiveness | -0.03 | **0.82** |
| Smooth flow | 0.05 | **0.79** |
| Equal participation | -0.15 | **0.65** |
| Engagement | 0.13 | **0.60** |
| Enjoyment | 0.64 | 0.32 |
| Agreement | 0.61 | 0.35 |
| Understanding | 0.58 | 0.35 |
| Trust | 0.56 | 0.37 |
| Comfortable with each other | 0.52 | 0.42 |
| Satisfaction | 0.44 | 0.50 |
| Harmony | 0.42 | 0.52 |
| Cooperation | 0.42 | 0.54 |

fatigue. The trimmed videos can be accessed with this Open Science Framework (OSF) link[2].

*3) Materials (Scale Items):* Participants were given the prompt "Rate how much you think the following was present in the interaction" to rate each of the 27 items listed in Table II on a five-point Likert scale ranging from 1 (*Strongly Disagree*) to 5 (*Strongly Agree*).

*4) Procedure:* After participants consented to join this study, they were randomly assigned to watch one of the four videos. Following the video, participants were instructed to rate the 27 scale items, which were presented in a randomized order. Participants were also asked to complete demographic questions, including gender, ethnicity, and age. Participants took an average of 3.23 minutes ($SD$ = 2.10 mins) to complete the study and were compensated with $1.25 USD.

### B. Results

We conducted an exploratory factor analysis (EFA) to determine the number of factors in our scale and the relevant scale items within each factor. Since Very Simple Structure (VSS), Empirical Bayesian Information Criterion (BIC), and Parallel Analysis all suggested a highly interpretable two-dimensional structure, we ran EFA with two factors using promax rotation on participants' responses. Upon examining the EFA results, we kept scale items that had one loading above or equal to 0.6 and had no cross-loadings above 0.3 (suggested by [84], [85]). For the remaining scale items, we inspected their overall internal reliability using Cronbach's alpha and McDonald's omega ($\alpha$ = 0.96, $\omega_{total}$ = 0.97). Since it is common to name factors in EFA [86], we named the two factors using items' contextual meanings: Connection (13 items) and Coordination (6 items), as shown in Table II.

[2]https://osf.io/5ezga/?view_only=f857949cfe1e49dd8f75ce1aac206b9f

### C. Discussion

In Study 1, we created a scale with 19 items to evaluate the construct of rapport. Through EFA, we discovered that our scale was multidimensional and had two core factors, Connection (13 items) and Coordination (6 items). We thus named our scale *Connection-Coordination Rapport (CCR) Scale* and found the scale showed high reliability.

To calculate a score for rapport with our CCR scale, we propose the following: (1) average the ratings from the items in the Connection factor, (2) average the ratings from the items in the Coordination factor, and (3) average the values from (1) and (2) to obtain a single score. The values from (1) and (2) can also be examined independently if the researchers prefer to focus on only one factor.

## V. Study 2 (Scale Evaluation)

Our two goals in Study 2 were to evaluate our 19-item CCR scale from Study 1 with a different set of videos and compare its efficacy in measuring rapport with an alternative state-of-the-art scale. We selected the 11-item Rapport Scale 4 designed to measure rapport between human and virtual agents from Gratch et al. [33] because it builds on prior versions of the scale [32], [87]–[93], with Rapport Scale 1 from Gratch et al. [32] being the most cited rapport scale on Google Scholar. We also chose to compare with this scale because the items can be easily modified into a third-person perspective and they are more applicable to a wide range of scenarios as opposed to other human-robot rapport scales (e.g., [1], [19], [37]). For the rest of this paper, we will refer to the 11-item Rapport Scale 4 from Gratch et al. [33] as the Gratch Rapport Scale.

### A. Method

We conducted an online within-subjects study asking participants to watch a new set of four videos of human-robot interactions (see Figure 2) and rate our 19-item CCR scale and the Gratch Rapport Scale for each video. After watching and rating all four videos, we also asked participants to rank the videos from highest rapport to lowest rapport (cf. [50]).

*1) Participants:* We recruited 250 participants through the Prolific platform who were native English speakers. The sample size of the participants was also suggested by the 10:1 criteria [79]. 49 participants were removed because they failed one or more attention checks. From the remaining 201 participants, there were 125 White, 46 Black or African American, 12 Asian, 8 other ethnicity, 8 identified as two or more ethnicities, and 2 preferred not to disclose. The age of the participants ranged from 18 to 69 ($M$ = 36.11, $SD$ = 11.46). 105 of them identified as women, 89 as men, 5 as non-binary, and 2 as other gender.

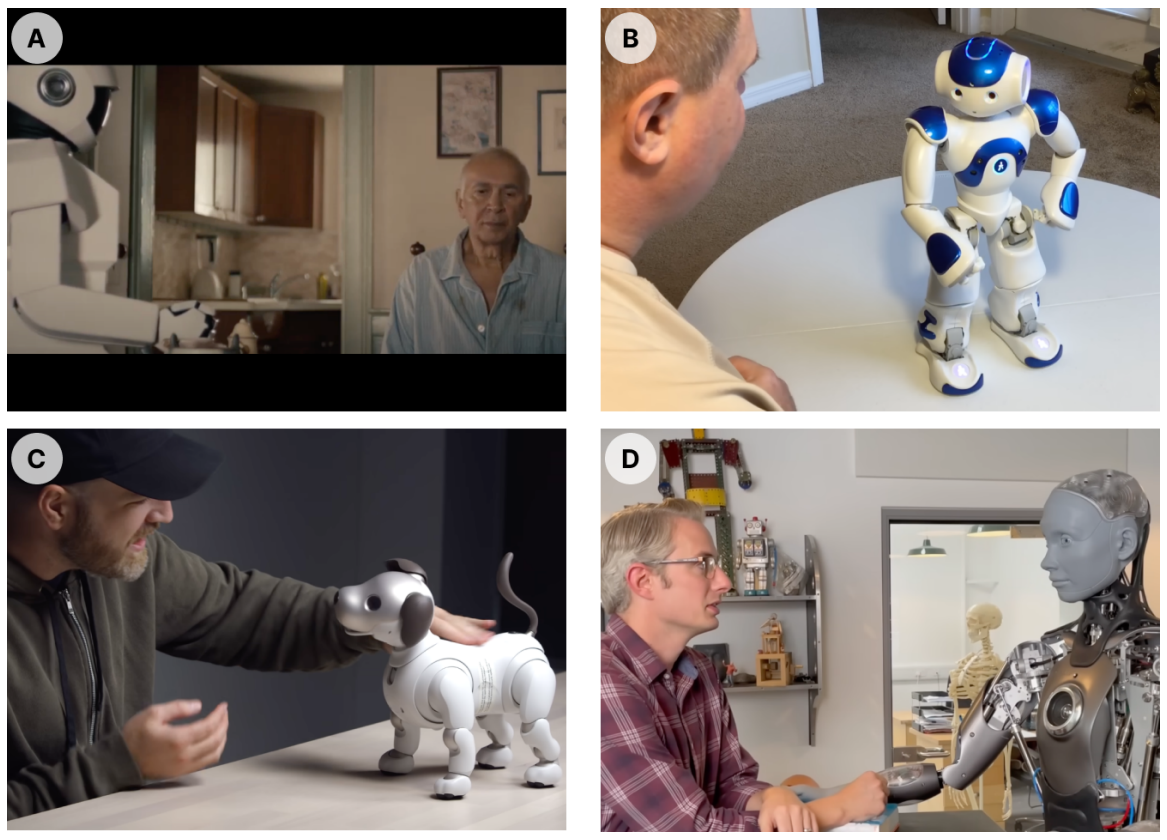

Fig. 2. Snapshots of the four videos (A) Healthcare, (B) Hold Hands, (C) Pet, and (D) Conversation used in Study 2.

*2) Materials (Videos):* As we did in Study 1, we selected another four YouTube videos [94]–[97] that were similarly involved interactions between a single robot and a single human but differed in the context of the situation from those used in Study 1. We labeled them as (a) *Healthcare* – a humanoid robot assists a dementia patient, (b) *Hold Hands* – a humanoid robot talks and holds hands with a person, (c) *Pet* – a dog-like robot acts as a person's pet companion, and (d) *Conversation* – a humanoid robot holds a conversation with a human. Snapshots of these videos are shown in Figure 2. These videos were trimmed to be under a minute to minimize participant boredom, and they can also be accessed with this OSF link [3].

*3) Materials (Scale Items):* We used the 19-item CCR scale developed from Study 1 (see the top 19 items listed in Table II) and the 11-item Gratch Rapport Scale [33] which was modified to be in the third person (i.e., we replaced all occurrences of "I" with "the person" and all occurrences of "the listener" with "the robot"). For example, the first scale item in the Gratch Rapport Scale was modified from "*I felt I had a connection with the listener*" to "*The person felt they had a connection with the robot*." In the modified Gratch Rapport Scale, participants were given the question "Please indicate the degree to which you felt each of the following conditions during the interaction between the person and the robot" to rate the statements on a five-point Likert scale ranging from 1 (*Strongly Disagree*) to 5 (*Strongly Agree*) (see Supplemental Materials Table SII for the modified Gratch Rapport Scale).

*4) Procedure:* After completing a consent form, participants were asked to watch four videos in a randomized order. After watching each video, participants rated the video using both our 19-item CCR scale and the modified Gratch Rapport Scale. When participants finished watching all four videos, they were provided with our definition of rapport (see Section II) and were asked to rank the four videos from lowest to highest rapport, using a scale with the following labels: *Very Low Rapport*, *Low Rapport*, *High Rapport*, and *Very High Rapport*. They were also asked to fill out demographic questions including gender, ethnicity, and age. Participants spent an average of

[3]https://osf.io/5ezga/?view_only=f857949cfe1e49dd8f75ce1aac206b9f

16.57 minutes ($SD$ = 7.44 mins) completing the survey and were compensated $3.00 USD.

### B. Results

Our overall goal was to evaluate our CCR scale via confirmatory factor analysis (CFA) and compare it with the current best available scale measuring rapport.

We first re-evaluated the scale items from our CCR scale within the context of the videos in Study 2 and found that the scale item "Deep conversation" was notably more context-specific than the other items (e.g., the Pet video in Figure 2C did not involve any verbal communication from the robot). We thus decided to remove "Deep conversation" from the CCR scale to make it more generalizable to a variety of robots and contexts (e.g., robots that communicate only non-verbally). For the rest of this section, we analyzed participants' ratings for our CCR scale without "Deep conversation."

We then conducted a CFA on participants' CCR scale ratings for the four videos to assess the underlying factor structure derived from EFA in Study 1. The CFA result showed that our CCR scale has a good fit based on Comparative Fit Index (CFI), Tucker-Lewis Index (TLI), and Standardized Root Mean Square Residual (SRMR): $CFI = 0.997$ ($\geq 0.95$), $TLI = 0.996$ ($\geq 0.95$), and $SRMR = 0.046$ ($\leq 0.08$). Meanwhile, the Root Mean Square Error of Approximation (RMSEA) only indicates a moderate fit [98]: $RMSEA = 0.096$ ($\leq 0.08$).

To compare our CCR scale with the Gratch Rapport Scale, we confirmed their strong correlation ($R = 0.84, p < 0.001$) to ensure they were both measuring the same construct of rapport and calculated their internal consistency with Cronbach's alpha and McDonald's omega. We found both scales were highly reliable: CCR scale ($\alpha = 0.97$, $\omega_{total} = 0.97$) and Gratch Rapport Scale ($\alpha = 0.89$, $\omega_{total} = 0.94$).

Next, we compared the model fit of an ordinal regression by using participants' video ranking of rapport as the dependent variable and the score of the scale as the independent variable to determine which scale is a better predictor of the raters' rank orderings (cf. [50]). We first found that both models are significantly better than chance ($p < 0.05$). We then evaluated how well each model fits the data by examining the Akaike Information Criterion (AIC) and found the model with our CCR scale has a significantly better model fit ($AIC = 1874.23$) compared to the model with the Gratch Rapport Scale ($AIC = 1914.33$). Note that if a model is more than 2 units of AIC lower than another model, it is considered a significantly better model [99]. We also assessed the model fit by Nagelkerke Pseudo $R_N^2$ and discovered the model with our CCR scale achieved a better model fit ($R_N^2 = 0.39$) than the one with the Gratch Rapport Scale ($R_N^2 = 0.35$). Both AIC and Nagelkerke Pseudo $R_N^2$ show that the ordinal regression model with our CCR scale is significantly preferred over the model with the Gratch Rapport Scale.

Lastly, we evaluated whether both scales could correctly identify the order of the four videos from lowest rapport to highest rapport according to the participant's average ranking (Healthcare < Hold Hands < Conversation < Pet). As shown

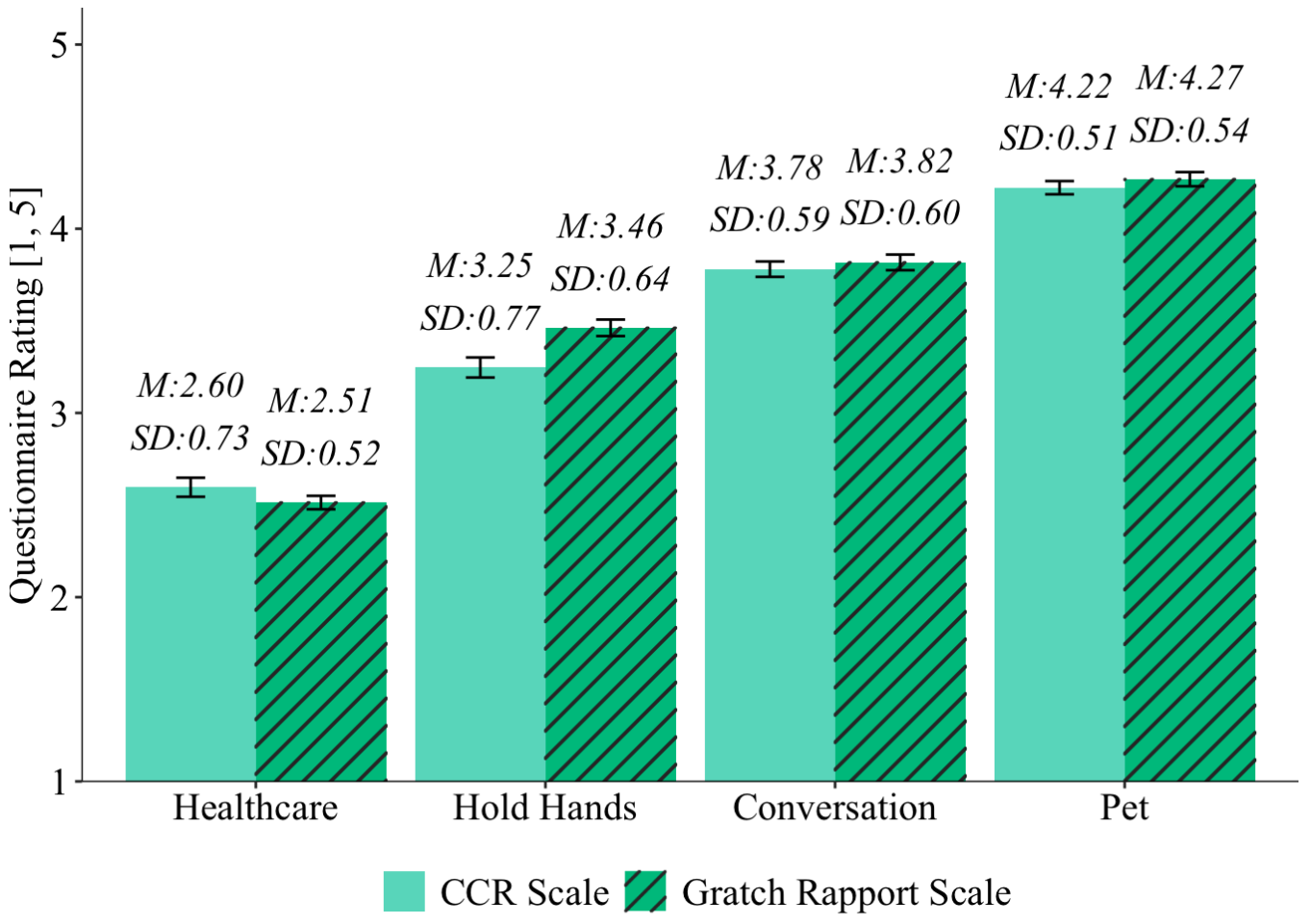

Fig. 3. CCR Scale and Gratch Rapport Scale could both accurately determine videos from the lowest rapport to the highest rapport, which is in line with the participant's average ranking (Healthcare < Hold Hands < Conversation < Pet). Error bars show one standard error from the mean.

in Figure 3, both scales accurately matched the participant's rapport ranking for the four videos.

### C. Discussion

We evaluated our CCR scale by asking a new group of participants to rate a new set of videos with robots that have different morphologies and behaviors. From the CFA results, we found our CCR scale has an acceptable fit. Our CCR scale also has excellent reliability as shown by Cronbach's alpha and McDonald's omega. Finally, our CCR scale showed a statistically better fit than the Gratch Rapport Scale.

Study 1 and Study 2 have shown that the CCR scale has strong psychometric properties to measure rapport from a third-person perspective by watching videos. However, the goal of many HRI researchers is to improve in-person human-robot interactions and assess the rapport a person senses between themselves and the robot from the first-person perspective. Thus, additional validation of the CCR scale was required to examine whether it can accurately measure rapport in in-person human-robot interactions from a first-person point of view.

## VI. Study 3 (Scale Validation)

The main objective of Study 3 was to validate our CCR scale from a first-person perspective in an in-person HRI study replicated from prior work. To determine which study to conduct, we reviewed past literature and found that perceived responsiveness could enhance well-being in relationships between humans [100], [101], humans and virtual agents [32], [59], [102], [103], and humans and robots [4], [104], [105]. In Birnbaum et al. [104], a robot's responsive actions (e.g., head-nodding and displaying personalized text on its screen) increased the robot's perceived sociability and competence. Additionally, Birnbaum et al. [104] showed the robot's responsiveness raised participants' desire for the robot's companion and the frequency of their approaching behaviors towards the robot (e.g., shortening physical proximity to the robot, leaning toward the robot, and maintaining eye contact with the robot). Given that other prior works also showed that human-robot rapport could be correlated with the robot's head-nodding gestures [16], [17], [28] and people's approaching behaviors towards the robot [2], [17], we chose to replicate the first study of the two studies conducted in Birnbaum et al. [104] to investigate the correlation between the robot's perceived responsiveness and rapport developed during the interaction. We hypothesized that people would feel stronger rapport when interacting with a more responsive robot than a less responsive one.

## A. Method

We replicated the study method in Birnbaum et al. [104] by conducting an in-person between-subjects study where participants interacted with either a responsive robot (**responsive condition**) or an unresponsive robot (**unresponsive condition**). In the study, participants disclosed a current problem, concern, or stressor across three messages to the robot. In the first message, participants talked about the circumstances of the concern. In the second message, they discussed their feelings and thoughts about the concern. In the third message, participants shared more details that are crucial to the concern.

In the **responsive condition**, the robot would nod its head and say one template-based sentence that was personalized and positive to the participant after they finished giving their response for each message. For example, if a participant said they had hesitation to relocate to a new city for a higher-paying job, the robot would nod and say "it sounds like you are facing a very tough decision that can be challenging to navigate." In the **unresponsive condition**, the robot would not perform any gestures, instead simply giving a one-sentence utterance that asked participants to move on when they completed their response for each message (e.g., "please go on to the next part"). The robot's response timing and content were controlled by a Wizard-of-Oz operator. In the **responsive condition**, the Wizard-of-Oz operator could also customize the robot's utterance. We show the message bank for the robot's possible responses in Supplemental Materials Table SIII. When the interaction concluded, participants completed a questionnaire that contained our 18-item CCR scale.

Since we did not have access to the non-humanoid robot (Travis) used in Birnbaum et al. [104], we used the Misty II robot instead. The Misty robot used both head nodding gestures and verbal language to deliver responses to participants, whereas the non-humanoid Travis robot in Birnbaum et al. used nodding gestures and displayed its responses as text on a screen. Further, we placed a laptop next to the Misty robot to remind participants of the prompt for each message (see Figure 4 for the study setup), while Birnbaum et al. did not include one in their setup. Despite these minor differences in the study protocol, the robot's responsiveness remained as similar as possible – both robots provided nonverbal head nodding and provided the same content in their text and verbal responses.

*1) Participants:* We determined the sample size of participants a priori using a power analysis with a power of 0.8 and an effect size $d$ of 0.8 at $p < 0.05$, which suggested 21 participants for each of the two conditions. We recruited a total of 44 participants at Mindworks, a behavioral science museum in downtown Chicago. 16 participants were White, 5 were Black or African American, 15 were Asian, 5 were other ethnicity, and 3 were identified as two or more ethnicities. Participants ranged in age from 18 to 79 ($M$ = 34.41, $SD$ = 15.46), and 26 of them identified as women, 16 as men, and 2 as non-binary.

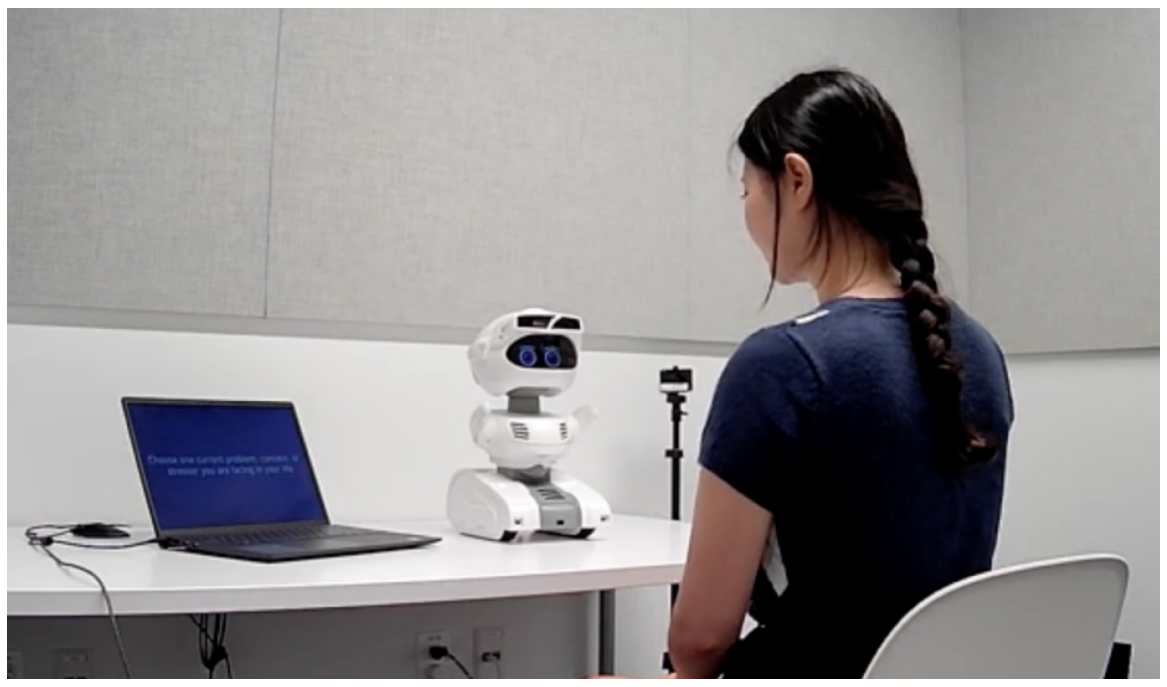

Fig. 4. In Study 3, a participant in the unresponsive condition disclosed a personal concern to the robot.

*2) Materials (Scale Items):* We incorporated our 18-item CCR scale shown in Table I to measure participants' perceived rapport with the robot. We also used the scale items from Birnbaum et at. [104] that measure perceived robot responsiveness, sociability, competence, and attractiveness, as well as participants' desire for robot's companionship (refer to Supplemental Materials Table SIV for the full list of questionnaire items).

*3) Procedure:* After the participant provided their demographic data and consented to participate in the study, the experimenter instructed the participant to sit on a chair facing the robot. The experimenter told the participant a cover story that we were testing a newly developed speech-comprehension algorithm for the robot and asked them to disclose a current problem, concern, or stressor to the robot by separating it into three messages. Participants were asked by the experimenter to say the statement "and that's it" when they finished saying each message to signal the robot that they were done speaking. In reality, this was the signal for the Wizard-of-Oz operator to control the robot's gestures and speech according to the designated condition. The experimenter then left the room and participants were given a maximum of 3 minutes per message to talk about their problem. Following the interaction, participants completed a questionnaire and were debriefed to ensure they were comfortable with their participation. Participants took an average of 16.42 minutes ($SD$ = 7.13 mins) to complete the study and were compensated with points equivalent to $4.00 USD that could be redeemed for gifts at Mindworks. During the study, we recorded the participant's interaction with the robot using two cameras.

## B. Results

We first calculated the correlation coefficient between participants' ratings of rapport from our CCR scale and their

TABLE III

Means, Standard Deviations, T-test Statistics, P Values, and Effect Sizes of the Measures in Study 3

| | Responsive Robot M | Responsive Robot SD | Unresponsive Robot M | Unresponsive Robot SD | $t_{(42)}$ | $p$ | Cohen's $d$ | 95% CI for $d$ |
|---|---|---|---|---|---|---|---|---|
| **Connection-Coordination Rapport (CCR) Scale** | 3.44 | 0.61 | 2.37 | 0.68 | 5.44 | $< 0.001$*** | 1.61 | (0.92, 2.30) |
| **Connection Factor in CCR Scale** | 3.36 | 0.68 | 1.95 | 0.71 | 6.76 | $< 0.001$*** | 2.00 | (1.27, 2.73) |
| **Coordination Factor in CCR Scale** | 3.52 | 0.67 | 2.80 | 0.80 | 3.25 | 0.002** | 0.96 | (0.33, 1.59) |
| Perceived Robot Responsiveness | 3.46 | 0.93 | 1.55 | 0.55 | 8.31 | $< 0.001$*** | 2.46 | (1.67, 3.25) |
| Perceived Robot Sociability | 4.73 | 1.22 | 2.26 | 1.00 | 7.34 | $< 0.001$*** | 2.17 | (1.42, 2.93) |
| Perceived Robot Competence | 4.03 | 1.46 | 2.32 | 1.27 | 4.16 | $< 0.001$*** | 1.23 | (0.58, 1.88) |
| Perceived Robot Attractiveness | 2.55 | 0.81 | 1.86 | 0.88 | 2.67 | 0.011* | 0.79 | (0.17, 1.41) |
| Participant's Desire for Robot Companionship | 2.09 | 1.09 | 1.21 | 0.37 | 3.62 | $< 0.001$*** | 1.07 | (0.43, 1.71) |
| Participant's Approach Behaviors | 3.57 | 1.00 | 2.14 | 0.83 | 5.15 | $< 0.001$*** | 1.52 | (0.85, 2.20) |
| Participant's Self-disclosure | 3.21 | 1.33 | 2.82 | 1.25 | 0.99 | 0.327 | 0.29 | (-0.31, 0.89) |

(*), (**), (***) denote $p < 0.05$, $p < 0.01$, and $p < 0.001$, respectively.

ratings of perceived robot responsiveness. As predicted, we confirmed that rapport and the robot's responsiveness have a fairly strong correlation ($R = 0.75, p < 0.001$). We then used t-tests to analyze our data between the two conditions by replicating the approach of statistical analysis in Birnbaum et al. [104]. The results are summarized in Table III.

*1) Connection-Coordination Rapport (CCR) Scale:* From participants' ratings of the 18 scale items in our CCR scale, we found high internal consistency ($\alpha = 0.95$, $\omega_{total} = 0.96$). We calculated the score of our CCR scale and found that, as predicted, participants in the responsive condition perceived significantly stronger rapport compared to those in the unresponsive condition (see Figure 5 and Table III).

We further examined participants' ratings of the Connection and Coordination factors between the two conditions. Participants in the responsive condition rated the robot significantly higher for both the Connection factor and the Coordination factor than participants in the unresponsive condition (see Figure 5 and Table III).

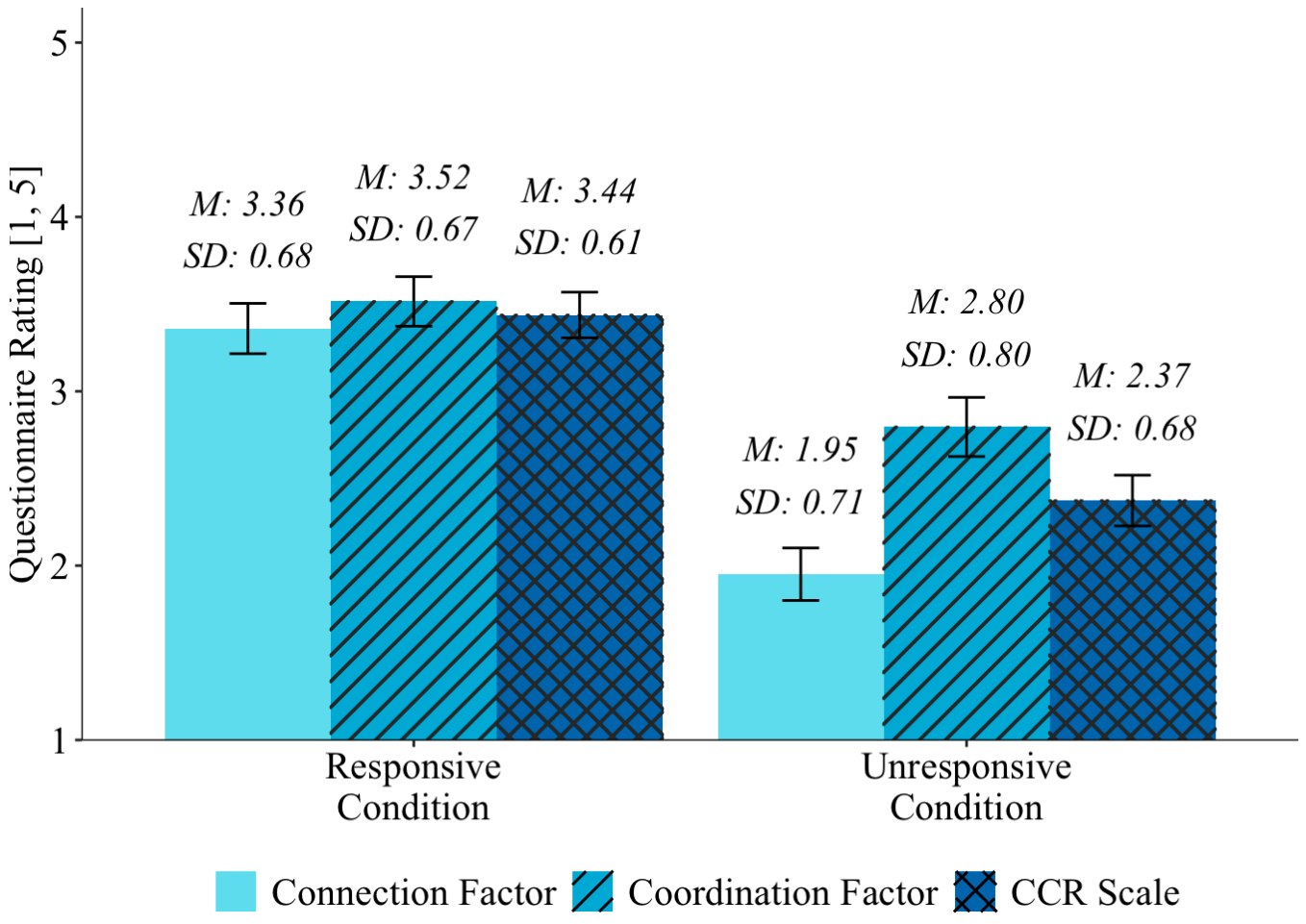

Fig. 5. Participants' responses for the Connection Factor, the Coordination Factor, and the CCR Scale between the responsive condition and the unresponsive condition. Error bars show one standard error from the mean.

*2) Perceived Robot Responsiveness, Sociability, Competence, and Attractiveness:* We averaged participants' responses to the questionnaire items measuring perceived robot responsiveness ($\alpha = 0.96$), sociability ($\alpha = 0.91$), competence ($\alpha = 0.93$), and attractiveness ($\alpha = 0.68$). We found that participants in the responsive condition perceived significantly higher responsiveness, sociability, competence, and attractiveness from the robot compared to those in the unresponsive condition (see Table III for the statistics). These significant findings replicated the results from Birnbaum et al. [104] except that participants in our study perceived the responsive robot as significantly more attractive than the unresponsive robot, where Birnbaum et al. observed no significant difference.

*3) Participant's Desire for Robot Companionship:* We took the average of the scale items that measure the participant's desire for companionship by the robot ($\alpha = 0.84$). Participants showed significantly more desire for the robot's companionship when they were in the responsive condition as opposed to the unresponsive condition (see Table III). This successfully replicated the finding in Birnbaum et al. [104].

*4) Participant's Approach Behaviors:* We had three coders watch the videos and rate an overall score ranging from 1 to 5. A rating of 1 indicates that the participant displayed approaching behaviors (e.g., shortening physical proximity, leaning toward the robot, smiling, and maintaining eye contact) for less than 20% of the time, and a rating of 5 indicates the participant showed those behaviors for more than 80% of the time. We calculated Krippendorff's Alpha for inter-rater reliability by asking three coders to evaluate five videos ($\alpha = 0.88$) and then had two coders rate all videos. We averaged the two coders' ratings for each video and found participants in the responsive condition showed significantly more approaching behaviors to the robot compared to those in the unresponsive condition (see Table III), successfully replicating the result in Birnbaum et al. [104].

*5) Participant's Self-disclosure:* We also had three coders watch the recorded videos and rate each of them on a scale from 1 to 5. A rating of 1 denotes participant's self-disclosure was very brief (e.g., spoke in one or two sentences) and lacked significant emotional depth, whereas a rating of 5 denotes participant's self-disclosure was extensive (e.g., spoke in paragraphs) and provided deep emotional or personal insights. Three coders rated an overlapping set of five videos ($\alpha$ = 0.91) and then two of them evaluated all videos. We took the average of the two coders' ratings for each video and found no significant differences between the conditions (see Table III). This non-significant result again replicated the finding in Birnbaum et al. [104].

### C. Discussion

The most important finding in Study 3 was that participants in the responsive condition perceived significantly greater rapport (as measured by our CCR scale) compared to those in the unresponsive condition. This significant difference between participants' ratings of rapport confirmed the construct validity of our CCR scale and gave support to our hypothesis that people establish deeper rapport with a robot when it is more responsive as opposed to less responsive.

As we further analyzed participants' ratings for the Connection factor and the Coordination factor in our CCR scale, we show in Figure 5 that the factors had similarly high ratings in the responsive condition, but they had markedly different ratings in the unresponsive condition where the Connection factor is rated lower than Coordination factor. This observation shows that while the absence of gestures and personalized speech from the unresponsive robot had a negative impact on both the Connection and Coordination factors, it had a greater influence on lowering the ratings in the Connection factor than the Coordination factor. We postulate that the ratings for the Coordination factor did not drop as much due to the nature of the study design where the robot would respond immediately after the participants said "and that's it", making them think this was a relatively coordinated turn-taking interaction. If the unresponsive robot's response is delayed or it speaks out of turn, we believe our CCR scale would still be able to capture the construct of rapport by yielding even lower ratings.

## VII. General Discussion

In this paper, we constructed the Connection-Coordination Rapport (CCR) Scale that effectively measures human-robot rapport. To create the CCR scale, we generated scale items using a mix of top-down approaches (cf. [50], [106], [107]) and bottom-up approaches (i.e., by soliciting definitions of rapport from online participants). We found the CCR scale contains two underlying factors through EFA in Study 1 and confirmed it has an overall good fit by CFA in Study 2, both of which asked participants to watch videos of human-robot interactions and rate them with our rapport-related items. We also found that our CCR scale showed excellent internal reliability and performed better than the Gratch Rapport Scale [33], as shown by the model fit comparison of an ordinal regression based on the participant's rapport ranking of the videos. In Study 3, we validated our CCR scale by replicating a prior HRI study (Birnbaum et al. [104]), showing the expected significant difference in participants' rating of rapport when they talked to a responsive robot as opposed to an unresponsive one. It is important to remember that the CCR scale is rigorously validated from both the third-person perspective (Studies 1 and 2) as well as the first-person perspective (Study 3).

As shown in Table I, the final product of our 18-item CCR scale contains two factors (Connection and Coordination), which align cohesively with the definition of rapport proposed in the prior psychology literature, Tickle-Degnen and Rosenthal [51]. Tickle-Degnen and Rosenthal proposed that rapport has three key components: mutual attentiveness, positivity, and coordination. In the CCR scale, we found participants mentally conceptualized scale items that describe mutual attentiveness and coordination (e.g., "Attentiveness", "Coordination") into one Coordination factor and considered items related to positivity (e.g., "Positivity", "Liking each other") as its own factor. The consistency of the conceptualization of rapport from Tickle-Degnen and Rosenthal [51] and the two dimensions of our CCR scale, therefore, provide additional validity and theoretical meaning to our scale's ability to measure rapport.

Our 18-item CCR scale has sufficiently addressed the weaknesses in past HRI works attempting to measure human-robot rapport. Improved upon past rapport scales (e.g., [19], [32], [33], [37]), we have made the items in the CCR scale brief and to the point, forward-coded, and applicable in a wide range of human-robot interaction scenarios. Further, similar to how trust scales in HRI are often designed to measure trust from a third-person perspective [108], [109], our scale can be directly used for evaluation of human-robot relationships from both first-person and third-person perspectives. Future HRI researchers can administer the CCR scale by having participants rate their relationship with a robot from a first-person perspective. Additionally, a robot designer could use the CCR scale to evaluate robot designs through videos created using Wizard-of-Oz methods before making significant prototyping efforts.

Since our CCR scale has only been evaluated with human-robot pairs, future work can test the scale with other pairs (e.g., human-human, human-virtual agent, or robot-robot pairs) to make the scale more generalizable while assessing its effectiveness. Rapport is also a construct that can be established when an interaction involves more than two individuals. Hence, future work can also apply the CCR scale to measure group rapport and compare its efficacy between first-person and third-person perspectives in an in-person experiment. Furthermore, we only validated the use of our CCR scale in short-term human-robot interactions in this work. Future work must investigate the scale's applicability to measure longer-term rapport.

In this work, we present the Connection-Coordination Rapport (CCR) Scale, which has been psychometrically validated. We hope future HRI researchers can adopt our CCR scale to measure rapport between humans and robots in their research.

## Acknowledgment

We thank Dr. Guy Hoffman for helping us to replicate the study from Birnbaum et al. [104] for our Study 3. We also thank Guan Chen, Rachel Liu, and staff members at Mindworks for their research assistance. This work was supported by the National Science Foundation (awards #2339581 and #2312352).

## References

[1] H. Kim, K. K. F. So, and J. Wirtz, "Service robots: Applying social exchange theory to better understand human–robot interactions," *Tourism Management*, vol. 92, p. 104537, 2022.

[2] M. K. Lee, J. Forlizzi, S. Kiesler, P. Rybski, J. Antanitis, and S. Savetsila, "Personalization in hri: A longitudinal field experiment," in *Proceedings of the seventh annual ACM/IEEE international conference on Human-Robot Interaction*, 2012, pp. 319–326.

[3] B. Song, M. Zhang, and P. Wu, "Driven by technology or sociality? use intention of service robots in hospitality from the human–robot interaction perspective," *International Journal of Hospitality Management*, vol. 106, p. 103278, 2022.

[4] G. Collins, "Improving human–robot interactions in hospitality settings," *International Hospitality Review*, vol. ahead-of-print, 04 2020.

[5] H. Qiu, M. Li, B. Shu, and B. Bai, "Enhancing hospitality experience with service robots: The mediating role of rapport building," *Journal of Hospitality Marketing & Management*, vol. 29, no. 3, pp. 247–268, 2020.

[6] J. M. Kory-Westlund, "Implications of children's social, emotional, and relational interactions with robots for human–robot empathy," *Conversations on Empathy: Interdisciplinary Perspectives on Empathy, Imagination and Othering*, pp. 256–279, 2023.

[7] N. Lubold, E. Walker, H. Pon-Barry, and A. Ogan, "Comfort with robots influences rapport with a social, entraining teachable robot," in *Artificial Intelligence in Education: 20th International Conference, AIED 2019, Chicago, IL, USA, June 25-29, 2019, Proceedings, Part I 20*. Springer, 2019, pp. 231–243.

[8] X. Tian, N. Lubold, L. Friedman, and E. Walker, "Understanding rapport over multiple sessions with a social, teachable robot," in *Artificial Intelligence in Education: 21st International Conference, AIED 2020, Ifrane, Morocco, July 6–10, 2020, Proceedings, Part II 21*. Springer, 2020, pp. 318–323.

[9] J. M. Kory-Westlund and C. Breazeal, "A long-term study of young children's rapport, social emulation, and language learning with a peer-like robot playmate in preschool," *Frontiers in Robotics and AI*, vol. 6, p. 81, 2019.

[10] D. McColl and G. Nejat, "Meal-time with a socially assistive robot and older adults at a long-term care facility," *Journal of Human-Robot Interaction*, vol. 2, no. 1, pp. 152–171, 2013.

[11] R. H. Wang, A. Sudhama, M. Begum, R. Huq, and A. Mihailidis, "Robots to assist daily activities: views of older adults with alzheimer's disease and their caregivers," *International psychogeriatrics*, vol. 29, no. 1, pp. 67–79, 2017.

[12] A. Vercelli, I. Rainero, L. Ciferri, M. Boido, and F. Pirri, "Robots in elderly care," *DigitCult-Scientific Journal on Digital Cultures*, vol. 2, no. 2, pp. 37–50, 2018.

[13] D. Hebesberger, T. Koertner, C. Gisinger, and J. Pripfl, "A long-term autonomous robot at a care hospital: A mixed methods study on social acceptance and experiences of staff and older adults," *International Journal of Social Robotics*, vol. 9, no. 3, pp. 417–429, 2017.

[14] C.-A. Smarr, T. L. Mitzner, J. M. Beer, A. Prakash, T. L. Chen, C. C. Kemp, and W. A. Rogers, "Domestic robots for older adults: attitudes, preferences, and potential," *International journal of social robotics*, vol. 6, pp. 229–247, 2014.

[15] S. H. Seo, K. Griffin, J. E. Young, A. Bunt, S. Prentice, and V. Loureiro-Rodríguez, "Investigating people's rapport building and hindering behaviors when working with a collaborative robot," *International Journal of Social Robotics*, vol. 10, pp. 147–161, 2018.

[16] K. Pasternak, Z. Wu, U. Visser, and C. Lisetti, "Towards building rapport with a human support robot," in *Robot World Cup*. Springer, 2021, pp. 214–225.

[17] ——, "Let's be friends! a rapport-building 3d embodied conversational agent for the human support robot," *arXiv preprint arXiv:2103.04498*, 2021.

[18] A. Bellas, S. Perrin, B. Malone, K. Rogers, G. Lucas, E. Phillips, C. Tossell, and E. de Visser, "Rapport building with social robots as a method for improving mission debriefing in human-robot teams," in *2020 Systems and Information Engineering Design Symposium (SIEDS)*. IEEE, 2020, pp. 160–163.

[19] T. Nomura and T. Kanda, "Rapport–expectation with a robot scale," *International Journal of Social Robotics*, vol. 8, pp. 21–30, 2016.

[20] R. Kirby, J. Forlizzi, and R. Simmons, "Affective social robots," *Robotics and Autonomous Systems*, vol. 58, no. 3, pp. 322–332, 2010.

[21] R. Gockley, A. Bruce, J. Forlizzi, M. Michalowski, A. Mundell, S. Rosenthal, B. Sellner, R. Simmons, K. Snipes, A. C. Schultz *et al.*, "Designing robots for long-term social interaction," in *2005 IEEE/RSJ International Conference on Intelligent Robots and Systems*. IEEE, 2005, pp. 1338–1343.

[22] J. Brixey and D. Novick, "Building rapport with extraverted and introverted agents," in *Advanced Social Interaction with Agents: 8th International Workshop on Spoken Dialog Systems*. Springer, 2019, pp. 3–13.

[23] T. W. Bickmore and R. W. Picard, "Establishing and maintaining long-term human-computer relationships," *ACM Transactions on Computer-Human Interaction (TOCHI)*, vol. 12, no. 2, pp. 293–327, 2005.

[24] J. M. Cortina, Z. Sheng, S. K. Keener, K. R. Keeler, L. K. Grubb, N. Schmitt, S. Tonidandel, K. M. Summerville, E. D. Heggestad, and G. C. Banks, "From alpha to omega and beyond! a look at the past, present, and (possible) future of psychometric soundness in the journal of applied psychology." *Journal of Applied Psychology*, vol. 105, no. 12, p. 1351, 2020.

[25] W. Revelle and R. E. Zinbarg, "Coefficients alpha, beta, omega, and the glb: Comments on sijtsma," *Psychometrika*, vol. 74, pp. 145–154, 2009.

[26] G. O. Boateng, T. B. Neilands, E. A. Frongillo, H. R. Melgar-Quiñonez, and S. L. Young, "Best practices for developing and validating scales for health, social, and behavioral research: a primer," *Frontiers in public health*, vol. 6, p. 149, 2018.

[27] D. B. McCoach, R. K. Gable, and J. P. Madura, *Instrument development in the affective domain*. Springer, 2013, vol. 10.

[28] L. D. Riek, P. C. Paul, and P. Robinson, "When my robot smiles at me: Enabling human-robot rapport via real-time head gesture mimicry," *Journal on Multimodal User Interfaces*, vol. 3, pp. 99–108, 2010.

[29] J. Suárez Álvarez, I. Pedrosa, L. M. Lozano, E. García Cueto, M. Cuesta Izquierdo, J. Muñiz Fernández *et al.*, "Using reversed items in likert scales: A questionable practice," *Psicothema, 30*, 2018.

[30] B. Weijters, H. Baumgartner, and N. Schillewaert, "Reversed item bias: an integrative model." *Psychological methods*, vol. 18, no. 3, p. 320, 2013.

[31] E. v. Sonderen, R. Sanderman, and J. C. Coyne, "Ineffectiveness of reverse wording of questionnaire items: Let's learn from cows in the rain," *PloS one*, vol. 8, no. 7, p. e68967, 2013.

[32] J. Gratch, N. Wang, J. Gerten, E. Fast, and R. Duffy, "Creating rapport with virtual agents," in *Intelligent Virtual Agents: 7th International Conference, IVA 2007 Paris, France, September 17-19, 2007 Proceedings 7*. Springer, 2007, pp. 125–138.

[33] J. Gratch, D. DeVault, G. M. Lucas, and S. Marsella, "Negotiation as a challenge problem for virtual humans," in *Intelligent Virtual Agents: 15th International Conference, IVA 2015, Delft, The Netherlands, August 26-28, 2015, Proceedings 15*. Springer, 2015, pp. 201–215.

[34] N. Lubold, E. Walker, and H. Pon-Barry, "Effects of adapting to user pitch on rapport perception, behavior, and state with a social robotic learning companion," *User Modeling and User-Adapted Interaction*, vol. 31, pp. 35–73, 2021.

[35] I. Buchem and N. Bäcker, "Nao robot as scrum master: results from a scenario-based study on building rapport with a humanoid robot in hybrid higher education settings," *Training, Education, and Learning Sciences*, vol. 59, pp. 65–73, 2022.

[36] R. Artstein, D. Traum, J. Boberg, A. Gainer, J. Gratch, E. Johnson, A. Leuski, and M. Nakano, "Listen to my body: Does making friends help influence people?" in *The Thirtieth International Flairs Conference*, 2017.

[37] Y. Kim and B. Mutlu, "How social distance shapes human–robot interaction," *International Journal of Human-Computer Studies*, vol. 72, no. 12, pp. 783–795, 2014.

[38] J. K. Fatima, M. I. Khan, S. Bahmannia, S. K. Chatrath, N. F. Dale, and R. Johns, "Rapport with a chatbot? the underlying role of anthropomorphism in socio-cognitive perceptions of rapport and e-word of mouth," *Journal of Retailing and Consumer Services*, vol. 77, p. 103666, 2024.

[39] T. Nomura and T. Kanda, "Differences of expectation of rapport with robots dependent on situations," in *Proceedings of the second international conference on Human-agent interaction*, 2014, pp. 383–389.

[40] E. Nichols, S. R. Siskind, L. Ivanchuk, G. Pérez, W. Kamino, S. Šabanović, and R. Gomez, "Hey haru, let's be friends! using the tiers of friendship to build rapport through small talk with the tabletop robot haru," in *2022 IEEE/RSJ International Conference on Intelligent Robots and Systems (IROS)*. IEEE, 2022, pp. 6101–6108.

[41] T. Nomura and T. Kanda, "Who expect rapport with robots? a survey-based study for analysis of people's expectation," *Tech. Rep.*, 2015.

[42] J. H. Wilson, R. G. Ryan, and J. L. Pugh, "Professor–student rapport scale predicts student outcomes," *Teaching of Psychology*, vol. 37, no. 4, pp. 246–251, 2010.

[43] M. D. Barnett, T. D. Parsons, and J. M. Moore, "Measuring rapport in neuropsychological assessment: the barnett rapport questionnaire," *Applied Neuropsychology: Adult*, vol. 28, no. 5, pp. 556–563, 2021.

[44] J. H. Wilson and R. G. Ryan, "Professor–student rapport scale: Six items predict student outcomes," *Teaching of Psychology*, vol. 40, no. 2, pp. 130–133, 2013.

[45] W. J. Lammers and J. A. Gillaspy Jr, "Brief measure of student-instructor rapport predicts student success in online courses," *International Journal for the Scholarship of Teaching and Learning*, vol. 7, no. 2, p. 16, 2013.

[46] R. G. Ryan, J. H. Wilson, and J. L. Pugh, "Psychometric characteristics of the professor–student rapport scale," *Teaching of Psychology*, vol. 38, no. 3, pp. 135–141, 2011.

[47] J. L. Schriver and R. Harr Kulynych, "Do professor–student rapport and mattering predict college student outcomes?" *Teaching of Psychology*, vol. 50, no. 4, pp. 342–349, 2023.

[48] C. D. White, K. S. Campbell, and K. M. Kacmar, "Development and validation of a measure of leader rapport management: the lrm scale," *Journal of Behavioral and Applied Management*, vol. 13, no. 2, pp. 121–149, 2012.

[49] R. Ryan and J. H. Wilson, "Professor-student rapport scale: Psychometric properties of the brief version," *Journal of the Scholarship of Teaching and Learning*, pp. 64–74, 2014.

[50] J. G. Trafton, J. M. McCurry, K. Zish, and C. R. Frazier, "The perception of agency," *ACM Transactions on Human-Robot Interaction*, vol. 13, no. 1, pp. 1–23, 2024.

[51] L. Tickle-Degnen and R. Rosenthal, "The nature of rapport and its nonverbal correlates," *Psychological inquiry*, vol. 1, no. 4, pp. 285–293, 1990.

[52] W. English, M. Gott, and J. Robinson, "The meaning of rapport for patients, families, and healthcare professionals: a scoping review," *Patient Education and Counseling*, vol. 105, no. 1, pp. 2–14, 2022.

[53] P. D. Koppel, K. P. Hye-young, L. S. Ledbetter, E. J. Wang, L. C. Rink, and J. C. De Gagne, "Rapport between nurses and adult patients with cancer in ambulatory oncology care settings: A scoping review," *International Journal of Nursing Studies*, p. 104611, 2023.

[54] M. J. Leach, "Rapport: A key to treatment success," *Complementary therapies in clinical practice*, vol. 11, no. 4, pp. 262–265, 2005.

[55] D. A. Neequaye and E. Mac Giolla, "The use of the term rapport in the investigative interviewing literature: A critical examination of definitions," *Meta-Psychology*, vol. 6, 2022.

[56] H. Spencer-Oatey, *Culturally speaking: Managing rapport through talk across cultures*. A&C Black, 2004.

[57] L. Huang, L.-P. Morency, and J. Gratch, "Virtual rapport 2.0," in *Intelligent Virtual Agents: 10th International Conference, IVA 2011, Reykjavik, Iceland, September 15-17, 2011. Proceedings 11*. Springer, 2011, pp. 68–79.

[58] D. D. Gremler and K. P. Gwinner, "Rapport-building behaviors used by retail employees," *Journal of Retailing*, vol. 84, no. 3, pp. 308–324, 2008.

[59] J. Gratch, M. Young, R. Aylett, D. Ballin, and P. Olivier, *Intelligent Virtual Agents: 6th International Conference, IVA 2006, Marina Del Rey, CA; USA, August 21-23, 2006, Proceedings*. Springer Science & Business Media, 2006, vol. 4133.

[60] H. Spencer-Oatey, "(im) politeness, face and perceptions of rapport: unpackaging their bases and interrelationships," *Journal of Politeness Research*, 2005.

[61] B. N. Frisby and M. M. Martin, "Instructor–student and student–student rapport in the classroom," *Communication Education*, vol. 59, no. 2, pp. 146–164, 2010.

[62] F. J. Bernieri, "Coordinated movement and rapport in teacher-student interactions," *Journal of Nonverbal behavior*, vol. 12, no. 2, pp. 120–138, 1988.

[63] D. D. Gremler and K. P. Gwinner, "Customer-employee rapport in service relationships," *Journal of service research*, vol. 3, no. 1, pp. 82–104, 2000.

[64] J. Duncombe and J. Jessop, "Doing rapport'and the ethics of 'faking friendship'," *Ethics in qualitative research*, vol. 2, 2002.

[65] L. K. Miles, L. K. Nind, and C. N. Macrae, "The rhythm of rapport: Interpersonal synchrony and social perception," *Journal of experimental social psychology*, vol. 45, no. 3, pp. 585–589, 2009.

[66] A. Abbe and S. E. Brandon, "The role of rapport in investigative interviewing: A review," *Journal of investigative psychology and offender profiling*, vol. 10, no. 3, pp. 237–249, 2013.

[67] F. J. Bernieri, J. S. Gillis, J. M. Davis, and J. E. Grahe, "Dyad rapport and the accuracy of its judgment across situations: a lens model analysis." *Journal of Personality and Social Psychology*, vol. 71, no. 1, p. 110, 1996.

[68] S. M. Miller, "The participant observer and over-rapport," *American Sociological Review*, vol. 17, no. 1, pp. 97–99, 1952.

[69] D. Walsh and R. Bull, "Examining rapport in investigative interviews with suspects: Does its building and maintenance work?" *Journal of police and criminal psychology*, vol. 27, pp. 73–84, 2012.

[70] J. Mumm and B. Mutlu, "Human-robot proxemics: physical and psychological distancing in human-robot interaction," in *Proceedings of the 6th international conference on Human-robot interaction*, 2011, pp. 331–338.

[71] J. Stolzenwald and P. Bremner, "Gesture mimicry in social human-robot interaction," in *2017 26th IEEE International Symposium on Robot and Human Interactive Communication (RO-MAN)*. IEEE, 2017, pp. 430–436.

[72] B. Irfan, A. Ramachandran, S. Spaulding, D. F. Glas, I. Leite, and K. L. Koay, "Personalization in long-term human-robot interaction," in *2019 14th ACM/IEEE International Conference on Human-Robot Interaction (HRI)*. IEEE, 2019, pp. 685–686.

[73] A. M. Aroyo, F. Rea, G. Sandini, and A. Sciutti, "Trust and social engineering in human robot interaction: Will a robot make you disclose sensitive information, conform to its recommendations or gamble?" *IEEE Robotics and Automation Letters*, vol. 3, no. 4, pp. 3701–3708, 2018.

[74] T. Yue, A. E. Janiw, A. Huus, S. Aguiñaga, M. Archer, K. Hoefle, and L. D. Riek, "Creating human-robot rapport with mobile sculpture," in *Proceedings of the seventh annual ACM/IEEE international conference on Human-Robot Interaction*, 2012, pp. 223–224.

[75] N. Mitsunaga, C. Smith, T. Kanda, H. Ishiguro, and N. Hagita, "Adapting robot behavior for human–robot interaction," *IEEE Transactions on Robotics*, vol. 24, no. 4, pp. 911–916, 2008.

[76] M. R. Fraune, B. C. Oisted, C. E. Sembrowski, K. A. Gates, M. M. Krupp, and S. Šabanović, "Effects of robot-human versus robot-robot behavior and entitativity on anthropomorphism and willingness to interact," *Computers in Human Behavior*, vol. 105, p. 106220, 2020.

[77] L. A. Fuente, H. Ierardi, M. Pilling, and N. T. Crook, "Influence of upper body pose mirroring in human-robot interaction," in *Social Robotics: 7th International Conference, ICSR 2015, Paris, France, October 26-30, 2015, Proceedings 7*. Springer, 2015, pp. 214–223.

[78] S. R. Brady, "Utilizing and adapting the delphi method for use in qualitative research," *International journal of qualitative methods*, vol. 14, no. 5, p. 1609406915621381, 2015.

[79] J. C. Nunnally, *Psychometric Theory*. McGraw-Hill, 1978.

[80] N. C. TV, "Ping pong playing robot," 2014, retrieved June 1, 2024. [Online]. Available: https://www.youtube.com/watch?v=7HfXd4aaElA

[81] L. W. I. From, "A pop-up japanese cafe with robot servers remotely controlled by people with disabilities," 2019, retrieved June 1, 2024; Timestamp: 1:55-2:19. [Online]. Available: https://www.youtube.com/watch?v=7HB6xLe2f3U

[82] M. R. by Embodied, "Meet moxie - the revolutionary robot companion for social-emotional learning," 2020, retrieved June 1, 2024; Timestamp: 0:13-1:00. [Online]. Available: https://www.youtube.com/watch?v=LQlNtxurleo

[83] A. Moon, "robot is mean," 2016, retrieved June 1, 2024; Timestamp: 0:00-0:40. [Online]. Available: https://www.youtube.com/watch?v=PAolTz5eBBs

[84] R. C. MacCallum, K. F. Widaman, S. Zhang, and S. Hong, "Sample size in factor analysis." *Psychological methods*, vol. 4, no. 1, p. 84, 1999.

[85] A. B. Costello and J. Osborne, "Best practices in exploratory factor analysis: Four recommendations for getting the most from your analysis," *Practical assessment, research, and evaluation*, vol. 10, no. 1, p. 7, 2019.

[86] M. W. Watkins, "Exploratory factor analysis: A guide to best practice," *Journal of black psychology*, vol. 44, no. 3, pp. 219–246, 2018.

[87] A. M. Von der Pütten, N. C. Krämer, J. Gratch, and S.-H. Kang, ""it doesn't matter what you are!" explaining social effects of agents and avatars," *Computers in Human Behavior*, vol. 26, no. 6, pp. 1641–1650, 2010.

[88] N. Wang and J. Gratch, "Can virtual human build rapport and promote learning?" in *Artificial Intelligence in Education*. IOS Press, 2009, pp. 737–739.

[89] S.-H. Kang and J. Gratch, "Socially anxious people reveal more personal information with virtual counselors that talk about themselves using intimate human back stories," *Annual Review of Cybertherapy and Telemedicine 2012*, pp. 202–206, 2012.

[90] D. M. Krum, S.-H. Kang, and M. Bolas, "Virtual coaches over mobile video," in *proceedings of the 27th International Conference on Computer Animation and Social Agents*, 2014.

[91] A. M. Von Der Pütten, N. C. Krämer, and J. Gratch, "How our personality shapes our interactions with virtual characters-implications for research and development," in *Intelligent Virtual Agents: 10th International Conference, IVA 2010, Philadelphia, PA, USA, September 20-22, 2010. Proceedings 10*. Springer, 2010, pp. 208–221.

[92] J. Appel, A. von der Pütten, N. C. Krämer, and J. Gratch, "Does humanity matter? analyzing the importance of social cues and perceived agency of a computer system for the emergence of social reactions during human-computer interaction," *Advances in Human-Computer Interaction*, vol. 2012, no. 1, p. 324694, 2012.

[93] B. Karacora, M. Dehghani, N. Kramer-Mertens, and J. Gratch, "The influence of virtual agents' gender and rapport on enhancing math performance," in *Proceedings of the Annual Meeting of the Cognitive Science Society*, vol. 34, no. 34, 2012.

[94] C. Datta, "Robot & frank 2012 breakfast scene," 2013, retrieved June 1, 2024; Timestamp: 0:04-0:57. [Online]. Available: https://www.youtube.com/watch?v=PKJcLnjky3s

[95] UAVAdvertising, "U.a.v advertising nao next gen robot," 2015, retrieved June 1, 2024; Timestamp: 7:10-8:02. [Online]. Available: https://www.youtube.com/watch?v=nbM0jplsTDw

[96] U. Therapy, "The $3000 sony aibo robot dog," 2018, retrieved June 1, 2024; Timestamp: 2:24-3:24. [Online]. Available: https://www.youtube.com/watch?v=8t8fyiiQVZ0

[97] E. Arts, "Ameca conversation using gpt 3 - will robots take over the world?" 2022, retrieved June 1, 2024; Timestamp: 0:01-0:57. [Online]. Available: https://www.youtube.com/watch?v=EWACmFLvpHE

[98] R. C. MacCallum, M. W. Browne, and H. M. Sugawara, "Power analysis and determination of sample size for covariance structure modeling." *Psychological methods*, vol. 1, no. 2, p. 130, 1996.

[99] T. W. Arnold, "Uninformative parameters and model selection using akaike's information criterion," *The Journal of Wildlife Management*, vol. 74, no. 6, pp. 1175–1178, 2010.

[100] H. T. Reis, "Responsiveness: Affective interdependence in close relationships." in *Mechanisms of social connection: From brain to group*. American Psychological Association, 2014, pp. 255–271.

[101] H. T. Reis and S. L. Gable, "Responsiveness," *Current Opinion in Psychology*, vol. 1, pp. 67–71, 2015.

[102] J. Gratch, A. Okhmatovskaia, F. Lamothe, S. Marsella, M. Morales, R. J. van der Werf, and L.-P. Morency, "Virtual rapport," in *Intelligent Virtual Agents: 6th International Conference, IVA 2006, Marina Del Rey, CA, USA, August 21-23, 2006. Proceedings 6*. Springer, 2006, pp. 14–27.

[103] J. Gratch, N. Wang, A. Okhmatovskaia, F. Lamothe, M. Morales, R. J. van der Werf, and L.-P. Morency, "Can virtual humans be more engaging than real ones?" in *Human-Computer Interaction. HCI Intelligent Multimodal Interaction Environments: 12th International Conference, HCI International 2007, Beijing, China, July 22-27, 2007, Proceedings, Part III 12*. Springer, 2007, pp. 286–297.

[104] G. E. Birnbaum, M. Mizrahi, G. Hoffman, H. T. Reis, E. J. Finkel, and O. Sass, "What robots can teach us about intimacy: The reassuring effects of robot responsiveness to human disclosure," *Computers in Human Behavior*, vol. 63, pp. 416–423, 2016.

[105] R. M. Stock and M. Merkle, "A service robot acceptance model: User acceptance of humanoid robots during service encounters," in *2017 IEEE international conference on pervasive computing and communications workshops (PerCom Workshops)*. IEEE, 2017, pp. 339–344.

[106] D. Ullman and B. F. Malle, "What does it mean to trust a robot? steps toward a multidimensional measure of trust," in *Companion of the 2018 acm/ieee international conference on human-robot interaction*, 2018, pp. 263–264.

[107] ——, "Measuring gains and losses in human-robot trust: Evidence for differentiable components of trust," in *2019 14th ACM/IEEE International Conference on Human-Robot Interaction (HRI)*. IEEE, 2019, pp. 618–619.

[108] ——, "Mdmt: multi-dimensional measure of trust," 2019.

[109] K. E. Schaefer, "Measuring trust in human robot interactions: Development of the "trust perception scale-hri"," in *Robust intelligence and trust in autonomous systems*. Springer, 2016, pp. 191–218.

# Supplemental Materials for Connection-Coordination Rapport (CCR) Scale: A Dual-Factor Scale to Measure Human-Robot Rapport

Ting-Han Lin[1], Hannah Dinner[2], Tsz Long Leung[1],
Bilge Mutlu[3], J. Gregory Trafton[4], and Sarah Sebo[1]

[1]University of Chicago, Chicago, USA. Emails: {tinghan, quincyleung, sarahsebo}@uchicago.edu

[2]University of Illinois Urbana-Champaign, Champaign, USA. Email: hdinner2@illinois.edu

[3]University of Wisconsin-Madison, Madison, USA. Email: bilge@cs.wisc.edu

[4]Naval Research Laboratory, Washington, USA. Email: greg.trafton@nrl.navy.mil

## I. Supplemental Materials

This PDF file includes Tables SI, SII, SIII, and SIV.

TABLE SI
SOURCES OF THE INITIAL POOL OF SCALE ITEMS AND REASONS FOR KEEPING OR REJECTING THEM BEFORE CONDUCTING STUDY 1

| **Scale Items** | **Dictionary Definitions** | **Google Scholar papers on "rapport"** | **Google Scholar papers on "rapport in human robot interaction"** | **Mentioned by Study 0 Participants (N=51)** | **Reasons for Keeping or Rejecting Them** |
|---|---|---|---|---|---|
| 1. Attentiveness | | [1]–[3] | [4]–[7] | N=1 (1.96%) | Kept |
| 2. Sympathy | [8]–[11] | | | N=1 (1.96%) | |
| 3. Empathy | [11], [12] | [13], [14] | | N=2 (3.92%) | |
| 4. Trust | [15], [16] | [17], [18] | [19]–[23] | N=7 (13.73%) | |
| 5. Warmth | | [18], [24] | [5] | N=2 (3.92%) | |
| 6. Excitement | | [2] | | N=0 (0.00%) | |
| 7. Enthusiasm | | [2] | [25] | N=0 (0.00%) | |
| 8. Positivity | | [1], [17] | [4]–[7] | N=7 (13.73%) | |
| 9. Smooth flow | | [26]–[29] | | N=3 (5.88%) | |
| 10. Understanding | [8], [10]–[12], [16], [30]–[32] | [14], [27], [33] | [20], [25], [34], [35] | N=13 (25.49%) | |
| 11. Harmony | [8], [9], [12] | [3], [18], [24], [26], [27], [29], [36], [37] | [5], [19], [25], [38] | N=3 (5.88%) | |
| 12. Agreement | [8], [12], [16] | | [6] | N=4 (7.84%) | |
| 13. Friendliness | [12] | [2], [14], [29] | [5], [25], [35], [38] | N=11 (21.57%) | |
| 14. Enjoyment | | [2], [14], [18], [24], [29] | [34], [35] | N=1 (1.96%) | |
| 15. Connection | [8], [9], [11] | [14], [24], [27], [33] | [21], [25], [34] | N=10 (19.61%) | |
| 16. Satisfaction | | [2] | [35] | N=0 (0.00%) | |
| 17. Cooperation | | [2], [29] | [38] | N=1 (1.96%) | |
| 18. Coordination | | [1], [27], [29] | [4]–[7], [38] | N=0 (0.00%) | |
| 19. Focus | | [29] | [5], [38] | N=0 (0.00%) | |
| 20. Engagement | | [27] | [25] | N=4 (7.84%) | |
| 21. Respect | | [3], [17] | [25] | N=6 (11.76%) | |
| 22. Liking each other | | [2], [29] | | N=0 (0.00%) | |
| 23. Closeness | [8], [10], [31] | [17], [18], [24] | [20]–[22] | N=2 (3.92%) | |
| 24. Equal participation | | [3] | | N=1 (1.96%) | |
| 25. Deep conversation | | [3], [14], [27], [39] | [34] | N=3 (5.88%) | |
| 26. Getting along | | | | N=4 (7.84%) | |
| 27. Comfortable with each other | | [18], [24], [29] | [5], [20] | N=8 (15.69%) | |
| 28. Being in sync | | [27], [39] | | N=0 (0.00%) | Rejected because they are idioms or too colloquial, which could be difficult to understand or translate into other languages |
| 29. Having chemistry | | | [19] | N=2 (3.92%) | |
| 30. A bond | [10] | [18], [24] | | N=2 (3.92%) | |
| 31. Being on the same wavelength | | | | N=1 (1.96%) | |
| 32. Clicking | | | [19] | N=1 (1.96%) | |
| 33. Mutuality | [12], [15], [16] | [1], [3], [13], [17], [18] | [4]–[6] | N=5 (9.80%) | Rejected because they are too broad or too vague |
| 34. Sharing a lot in common | | | [20] | N=2 (3.92%) | |
| 35. Good communication | [12] | [14], [17] | [25] | N=10 (19.61%) | |
| 36. Worthwhileness | | [29] | [38] | N=0 (0.00%) | |
| 37. Involvement | | [29] | [25], [38] | N=0 (0.00%) | |
| 38. Activeness | | [3], [29] | [23], [38] | N=2 (3.92%) | |
| 39. Interest in each other | | [2], [3], [14], [18], [24] | [5], [22], [25] | N=3 (5.88%) | |
| 40. Openness | | [3] | | N=6 (11.76%) | |
| 41. Self-disclosure | | | [20] | N=0 (0.00%) | Rejected because they are difficult to assess |
| 42. Coordinated postural movements | | | [5] | N=0 (0.00%) | |
| 43. Head nods | | | [5] | N=0 (0.00%) | |
| 44. Mutual gaze | | | [5] | N=0 (0.00%) | |
| 45. Initiating personal connection | | [18], [24] | | N=1 (1.96%) | |
| 46. Familiarity with each other | | | | N=4 (7.84%) | |
| 47. Being in a team | | | [19] | N=0 (0.00%) | Rejected because they are not relevant enough to rapport |
| 48. Motivation | | [2] | | N=0 (0.00%) | |
| 49. Humor | | [2], [14], [18], [24] | | N=0 (0.00%) | |
| 50. Naturalness | | | [5] | N=1 (1.96%) | |
| 51. Engrossment | | [29] | [38] | N=0 (0.00%) | |
| 52. Smile | | | [5] | N=0 (0.00%) | |
| 53. Giving advice | | [14] | [34] | N=0 (0.00%) | |
| 54. Sharing knowledge | | [14] | | N=2 (3.92%) | |
| 55. Asking questions | | [14] | | N=0 (0.00%) | |
| 56. Happiness | | [2] | | N=0 (0.00%) | Rejected because it is captured by excitement, enthusiasm, and positivity |
| 57. Smooth interaction | | [26], [28], [29] | | N=2 (3.92%) | Rejected because it is similar to smooth flow |
| 58. Frustration | | [2] | | N=1 (1.96%) | Rejected because they are reverse-coded scale items |
| 59. Anger | | [3] | | N=0 (0.00%) | |
| 60. Disgust | | [2] | | N=0 (0.00%) | |
| 61. Boredom | | [2], [27], [29] | | N=1 (1.96%) | |
| 62. Dullness | | [29] | | N=0 (0.00%) | |
| 63. Slowness | | [29] | | N=0 (0.00%) | |
| 64. Awkwardness | | [29] | | N=5 (9.80%) | |
| 65. Miscommunication | | | | N=2 (3.92%) | |
| 66. Surface level | | | | N=3 (5.88%) | |
| 67. Confusion | | | | N=2 (3.92%) | |

TABLE SII

MODIFIED RAPPORT SCALE 4 FROM GRATCH ET AL. USED IN STUDY 2

| Number | Scale Item |
|---|---|
| 1 | [The person] felt [they] had a connection with the [robot]. |
| 2 | [The person] think[s] that [they and the robot] understood each other. |
| 3 | The [robot] was warm and caring. |
| 4 | The [robot] was respectful to [the person]. |
| 5 | [The person] felt [they] had no connection with the [robot]. (reverse coded) |
| 6 | The [robot] created a sense of closeness or camaraderie between [them]. |
| 7 | The [robot] created a sense of distance between [them]. (reverse coded) |
| 8 | The [robot] communicated coldness rather than warmth. (reverse coded) |
| 9 | [The person] wanted to maintain a sense of distance between [them]. (reverse coded) |
| 10 | [The person] tried to create a sense of closeness or camaraderie between [them]. |
| 11 | [The person] tried to communicate coldness rather than warmth. (reverse coded) |

**Question wording:** Please indicate the degree to which you felt each of the following conditions during the interaction between the person and the robot. Please consider each question separately. **A five-point scale was used:** Strongly Disagree, Disagree, Neither Agree or Disagree, Agree, Strongly Agree. **Note:** Rapport Scale 4 is from Gratch et al. [40], and the texts placed in the squared brackets were modified.

TABLE SIII

WIZARD-OF-OZ MESSAGE BANK FOR THE ROBOT'S RESPONSE IN RESPONSIVE CONDITION AND UNRESPONSIVE CONDITION IN STUDY 3

| Number | Responsive Condition Message 1 |
|---|---|
| 1 | That's really tough. You must have gone through a very difficult time especially [a topic discussed by the participant] |
| 2 | I'm sorry to hear about [a topic discussed by the participant] |
| 3 | It sounds like you've been experiencing some tension in your relationship. That can be challenging to navigate |
| 4 | It sounds like a lot of pressure, especially given the expectations you may be facing |
| 5 | It sounds like you're experiencing a lot of change right now |
| **Number** | **Responsive Condition Message 2** |
| 1 | It's completely understandable that this situation would make you feel this way, given that [a topic discussed by the participant] |
| 2 | It's understandable to be concerned. Relationships are deeply personal and conflicts can really affect you |
| 3 | It must be tough to feel stressed and overwhelmed, especially when there's so much at stake |
| 4 | It must be incredibly disheartening to feel this way, especially when it matters so much to you |
| **Number** | **Responsive Condition Message 3** |
| 1 | What you're experiencing can be very difficult to navigate. I hope you get this resolved soon |
| 2 | It's clear that this situation is weighing heavily on you, but it seems like you are handling it well |
| 3 | It sounds like this is having a significant impact on your life |
| 4 | It sounds frustrating, but I'm sure you will be able to find a solution |
| 5 | It sounds like you gained some really important insights from this experience |
| **Number** | **Unresponsive Condition Message 1** |
| 1 | Please go on to the next part |
| **Number** | **Unresponsive Condition Message 2** |
| 1 | Please move on to the next message |
| **Number** | **Unresponsive Condition Message 3** |
| 1 | Please call the experimenter |

In the responsive condition, the Wizard-of-Oz operator would pick one of the speech responses and can customize it if necessary for each message. In the unresponsive condition, the Wizard-of-Oz operator only has one choice of response for each message and cannot customize it. These messages from both conditions were adapted from the first study of the two studies conducted in Birnbaum et al. [41].

TABLE SIV

PERCEIVED ROBOT RESPONSIVENESS, SOCIABILITY, COMPETENCE, AND ATTRACTIVENESS AND DESIRE FOR COMPANIONSHIP SURVEY USED IN STUDY 3

| Number | Perceived Robot Responsiveness Scale Item |
|---|---|
| 1 | Misty was responsive to what I said |
| 2 | Misty really listened to me |
| 3 | Misty understood me |
| 4 | Misty seemed interested in what I was thinking and feeling |
| 5 | Misty was on "the same wavelength" with me |
| 6 | Misty sees the "real" me |
| 7 | Misty was aware of what I was thinking and feeling |
| 8 | Misty was responsive to my needs |
| 9 | Misty expressed liking and encouragement for me |
| **Number** | **Perceived Robot Sociability Scale Item** |
| 1 | To what extent do you think that Misty is cooperative? |
| 2 | To what extent do you think that Misty is social? |
| 3 | To what extent do you think that Misty is friendly? |
| 4 | To what extent do you think that Misty is warm? |
| **Number** | **Perceived Robot Competence Scale Item** |
| 1 | To what extent do you think that Misty is knowledgeable? |
| 2 | To what extent do you think that Misty showed self-awareness? |
| 3 | To what extent do you think that Misty is competent? |
| 4 | To what extent do you think that Misty is responsible? |
| **Number** | **Perceived Robot Attractiveness Scale Item** |
| 1 | How attractive is Misty? |
| 2 | How hot is Misty? |
| 3 | How sophisticated is Misty? |
| 4 | How sexy is Misty? |
| 5 | How innovative is Misty? |
| 6 | How thought-provoking is Misty? |
| **Number** | **Desire for Companionship Scale Item** |
| 1 | To what extent do you want Misty to keep you company during stressful events, such as a dental treatment and or a difficult test? |
| 2 | To what extent do you want Misty to keep you company when you are alone? |

**Question wording**: Please rate the following statements. The perceived robot attractiveness survey used a seven-point scale from Not at all to Very Much and the rest of the surveys used a five-point scale from Not at all to Very Much. **Note:** These scale items were adapted from the first study of the two studies conducted in Birnbaum et al. [41]. Specifically, we changed the robot's name to Misty for all scale items.

# References

[1] L. Tickle-Degnen and R. Rosenthal, "The nature of rapport and its nonverbal correlates," *Psychological inquiry*, vol. 1, no. 4, pp. 285–293, 1990.

[2] F. J. Bernieri, "Coordinated movement and rapport in teacher-student interactions," *Journal of Nonverbal behavior*, vol. 12, no. 2, pp. 120–138, 1988.

[3] D. Walsh and R. Bull, "Examining rapport in investigative interviews with suspects: Does its building and maintenance work?" *Journal of police and criminal psychology*, vol. 27, pp. 73–84, 2012.

[4] K. Pasternak, Z. Wu, U. Visser, and C. Lisetti, "Towards building rapport with a human support robot," in *Robot World Cup*. Springer, 2021, pp. 214–225.

[5] ——, "Let's be friends! a rapport-building 3d embodied conversational agent for the human support robot," *arXiv preprint arXiv:2103.04498*, 2021.

[6] M. R. Fraune, B. C. Oisted, C. E. Sembrowski, K. A. Gates, M. M. Krupp, and S. Šabanović, "Effects of robot-human versus robot-robot behavior and entitativity on anthropomorphism and willingness to interact," *Computers in Human Behavior*, vol. 105, p. 106220, 2020.

[7] N. Lubold, E. Walker, H. Pon-Barry, and A. Ogan, "Comfort with robots influences rapport with a social, entraining teachable robot," in *Artificial Intelligence in Education: 20th International Conference, AIED 2019, Chicago, IL, USA, June 25-29, 2019, Proceedings, Part I 20*. Springer, 2019, pp. 231–243.

[8] "Collins english dictionary," 2024. [Online]. Available: https://www.collinsdictionary.com/

[9] "Dictionary.com," 2024. [Online]. Available: https://www.dictionary.com/

[10] "The chambers dictionary," 2024. [Online]. Available: https://chambers.co.uk/

[11] "Oxford english dictionary," 2024. [Online]. Available: https://www.oed.com/

[12] "The merriam-webster dictionary," 2024. [Online]. Available: https://www.merriam-webster.com/

[13] W. English, M. Gott, and J. Robinson, "The meaning of rapport for patients, families, and healthcare professionals: a scoping review," *Patient Education and Counseling*, vol. 105, no. 1, pp. 2–14, 2022.

[14] D. D. Gremler and K. P. Gwinner, "Rapport-building behaviors used by retail employees," *Journal of Retailing*, vol. 84, no. 3, pp. 308–324, 2008.

[15] "The american heritage dictionary of the english language," 2024. [Online]. Available: https://www.ahdictionary.com/

[16] "Vocabulary.com," 2024. [Online]. Available: https://www.vocabulary.com/

[17] D. A. Neequaye and E. Mac Giolla, "The use of the term rapport in the investigative interviewing literature: A critical examination of definitions," *Meta-Psychology*, vol. 6, 2022.

[18] B. N. Frisby and M. M. Martin, "Instructor–student and student–student rapport in the classroom," *Communication Education*, vol. 59, no. 2, pp. 146–164, 2010.

[19] S. H. Seo, K. Griffin, J. E. Young, A. Bunt, S. Prentice, and V. Loureiro-Rodríguez, "Investigating people's rapport building and hindering behaviors when working with a collaborative robot," *International Journal of Social Robotics*, vol. 10, pp. 147–161, 2018.

[20] Y. Kim and B. Mutlu, "How social distance shapes human–robot interaction," *International Journal of Human-Computer Studies*, vol. 72, no. 12, pp. 783–795, 2014.

[21] A. Bellas, S. Perrin, B. Malone, K. Rogers, G. Lucas, E. Phillips, C. Tossell, and E. de Visser, "Rapport building with social robots as a method for improving mission debriefing in human-robot teams," in *2020 Systems and Information Engineering Design Symposium (SIEDS)*. IEEE, 2020, pp. 160–163.

[22] H. Kim, K. K. F. So, and J. Wirtz, "Service robots: Applying social exchange theory to better understand human–robot interactions," *Tourism Management*, vol. 92, p. 104537, 2022.

[23] A. M. Aroyo, F. Rea, G. Sandini, and A. Sciutti, "Trust and social engineering in human robot interaction: Will a robot make you disclose sensitive information, conform to its recommendations or gamble?" *IEEE Robotics and Automation Letters*, vol. 3, no. 4, pp. 3701–3708, 2018.

[24] D. D. Gremler and K. P. Gwinner, "Customer-employee rapport in service relationships," *Journal of service research*, vol. 3, no. 1, pp. 82–104, 2000.

[25] N. Lubold, E. Walker, and H. Pon-Barry, "Effects of adapting to user pitch on rapport perception, behavior, and state with a social robotic learning companion," *User Modeling and User-Adapted Interaction*, vol. 31, pp. 35–73, 2021.

[26] H. Spencer-Oatey, "(im) politeness, face and perceptions of rapport: unpackaging their bases and interrelationships," *Journal of Politeness Research*, 2005.

[27] J. Gratch, N. Wang, J. Gerten, E. Fast, and R. Duffy, "Creating rapport with virtual agents," in *Intelligent Virtual Agents: 7th International Conference, IVA 2007 Paris, France, September 17-19, 2007 Proceedings 7*. Springer, 2007, pp. 125–138.

[28] A. Abbe and S. E. Brandon, "The role of rapport in investigative interviewing: A review," *Journal of investigative psychology and offender profiling*, vol. 10, no. 3, pp. 237–249, 2013.

[29] F. J. Bernieri, J. S. Gillis, J. M. Davis, and J. E. Grahe, "Dyad rapport and the accuracy of its judgment across situations: a lens model analysis." *Journal of Personality and Social Psychology*, vol. 71, no. 1, p. 110, 1996.

[30] "Cambridge dictionary," 2024. [Online]. Available: https://dictionary.cambridge.org/

[31] "Google's english dictionary (oxford languages)," 2024. [Online]. Available: https://www.google.com/

[32] "Oxford learner's dictionaries," 2024. [Online]. Available: https://www.oxfordlearnersdictionaries.com/

[33] J. Gratch, M. Young, R. Aylett, D. Ballin, and P. Olivier, *Intelligent Virtual Agents: 6th International Conference, IVA 2006, Marina Del Rey, CA; USA, August 21-23, 2006, Proceedings*. Springer Science & Business Media, 2006, vol. 4133.

[34] T. Nomura and T. Kanda, "Rapport–expectation with a robot scale," *International Journal of Social Robotics*, vol. 8, pp. 21–30, 2016.

[35] L. D. Riek, P. C. Paul, and P. Robinson, "When my robot smiles at me: Enabling human-robot rapport via real-time head gesture mimicry," *Journal on Multimodal User Interfaces*, vol. 3, pp. 99–108, 2010.

[36] M. J. Leach, "Rapport: A key to treatment success," *Complementary therapies in clinical practice*, vol. 11, no. 4, pp. 262–265, 2005.

[37] H. Spencer-Oatey, *Culturally speaking: Managing rapport through talk across cultures*. A&C Black, 2004.

[38] T. Yue, A. E. Janiw, A. Huus, S. Aguiñaga, M. Archer, K. Hoefle, and L. D. Riek, "Creating human-robot rapport with mobile sculpture," in *Proceedings of the seventh annual ACM/IEEE international conference on Human-Robot Interaction*, 2012, pp. 223–224.

[39] L. Huang, L.-P. Morency, and J. Gratch, "Virtual rapport 2.0," in *Intelligent Virtual Agents: 10th International Conference, IVA 2011, Reykjavik, Iceland, September 15-17, 2011. Proceedings 11*. Springer, 2011, pp. 68–79.

[40] J. Gratch, D. DeVault, G. M. Lucas, and S. Marsella, "Negotiation as a challenge problem for virtual humans," in *Intelligent Virtual Agents: 15th International Conference, IVA 2015, Delft, The Netherlands, August 26-28, 2015, Proceedings 15*. Springer, 2015, pp. 201–215.

[41] G. E. Birnbaum, M. Mizrahi, G. Hoffman, H. T. Reis, E. J. Finkel, and O. Sass, "What robots can teach us about intimacy: The reassuring effects of robot responsiveness to human disclosure," *Computers in Human Behavior*, vol. 63, pp. 416–423, 2016.